\documentclass[letterpaper, 10 pt, conference]{ieeeconf}  

\IEEEoverridecommandlockouts                              

\usepackage{makecell}
\usepackage{booktabs}
\usepackage{multirow}
\usepackage{hyperref}
\usepackage{url}
\usepackage{xcolor}
\usepackage{xcolor}
\definecolor{darkgreen}{RGB}{0,100,0}
\definecolor{darkred}{RGB}{139,0,0}
\usepackage{graphicx}
\usepackage{amsmath}
\usepackage{cleveref}
\crefname{section}{Sec.}{Secs.}

\newcommand{\eg}{e.g.}
\newcommand{\ie}{i.e.}

\Crefname{section}{Section}{Sections}
\Crefname{table}{Table}{Tables}
\crefname{table}{Tab.}{Tabs.}
\crefname{figure}{Fig.}{Figs.}
\crefname{algorithm}{Alg.}{Algs.}
\crefformat{equation}{Eq.~#2#1#3}
\usepackage[flushleft]{threeparttable}

\title{\LARGE \bf
Hi-Map: Hierarchical Factorized Radiance Field for High-Fidelity Monocular Dense Mapping}

 \author{Tongyan Hua$^{\dagger}$, Haotian Bai$^{\dagger}$, Zidong Cao, Ming Liu, Dacheng Tao, and Lin Wang* 
 \thanks{*Corresponding author. $^{\dagger}$Authors with equal contribution.}
  \thanks{T. Hua, H. Bai, and Z. Cao are with HKUST(GZ), China (t.hua.msc@outlook.com; haotianwhite@outlook.com; caozidong1996@gmail.com)}
 \thanks{M. Liu and L. Wang are with HKUST(GZ), Guangzhou and  HKUST, HongKong, China (eelium@ust.hk; linwang@ust.hk)}
 \thanks{D. Tao is with University of Sydney, Australia (dacheng.tao@sydney.edu.au)}
 }

\begin{document}

\maketitle

\thispagestyle{empty}
\pagestyle{empty}

\begin{abstract}
 
In this paper, we introduce \textbf{Hi-Map}, a novel monocular dense mapping approach based on Neural Radiance Field (NeRF). 
Hi-Map is exceptional in its capacity to achieve efficient and high-fidelity mapping using only posed RGB inputs. 
Our method eliminates the need for external depth priors derived from \eg, a depth estimation model. 
%
Our key idea is to represent the scene as a \underline{hierarchical} feature grid that encodes the radiance and then factorizes it into feature planes and vectors. As such, the scene representation becomes simpler and more generalizable for fast and smooth convergence on new observations. This allows for efficient computation while alleviating noise patterns by reducing the complexity of the scene representation.
%
Buttressed by the hierarchical factorized representation, we leverage the Sign Distance Field (SDF) as a proxy of rendering for inferring the volume density, demonstrating high mapping fidelity.
%
Moreover, we introduce a dual-path encoding strategy to strengthen the photometric cues and further boost the mapping quality, especially for the distant and textureless regions.
Extensive experiments demonstrate our method's superiority in geometric and textural accuracy over the state-of-the-art NeRF-based monocular mapping methods.

\end{abstract}


\section{INTRODUCTION}

Building high-fidelity dense 3D maps is essential for embodied intelligent systems, such as robots. 
The 3D maps enable the robots to perform scene-understanding tasks and navigate within complex and dynamic environments. As a result, timely feedback can be provided to humans, allowing them to control the robots through seamless physical interaction~\cite{human-robot1,human-robot2}.
Traditional dense mapping techniques, \eg, ~\cite{tradition1,tradition2,tradition3,tradition4} struggle to balance memory efficiency with accuracy. These methods often rely on explicitly tracking and storing co-observed points, which are later transformed into, for instance, the occupancy grid~\cite{occ14} or TSDF~\cite{RGBDSLAM5,RGBDSLAM6,occ-sdf18} to represent the scene. Consequently, The larger the number of points that are correctly tracked, the higher the fidelity of the map that can be generated, but this also requires a considerable amount of computation and storage.

With the advent of Neural Radiance Fields (NeRF)~\cite{nerf21}, several research attempts~\cite{imap22,niceslam22,voxfusion22,eslam23,coslam23} leverage neural field to better represent the scene by encoding the appearance and geometry in a compact and learnable way, benefiting both memory consumption and mapping quality.
NeRF-based dense mapping methods predominantly depend on input depth priors to facilitate online convergence by narrowing the search scope for sampling. Such depth priors usually derive from sensors~\cite{H2Mapping23,rim23,shine-map23,sbumap_mips23,tof23}. Alternatively, the depth estimation is provided by monocular visual Simultaneous Localization And Mapping (vSLAM) systems~\cite{nerfslam22,orbeez23,goslam23,newton23,hislam23arXiv} or depth estimation models~\cite{nicer23,hislam23arXiv}. 
However, this reliance on depth priors becomes a hurdle in resource-limited environments or situations where depth cues are either unavailable or unreliable. Even though the depth estimation can be internalized by adding the warping constraint when optimizing implicit representations~\cite{dimslam23}, it still struggles to achieve a balance between accuracy and computational efficiency.
Therefore, it is meaningful to achieve efficient and high-fidelity dense mapping without reliance on depth priors. This demands that the NeRF efficiently and swiftly generalizes to new observations where the underlying geometry is unknown.
\begin{figure}[t!]
      \centering
    \includegraphics[scale=0.34]{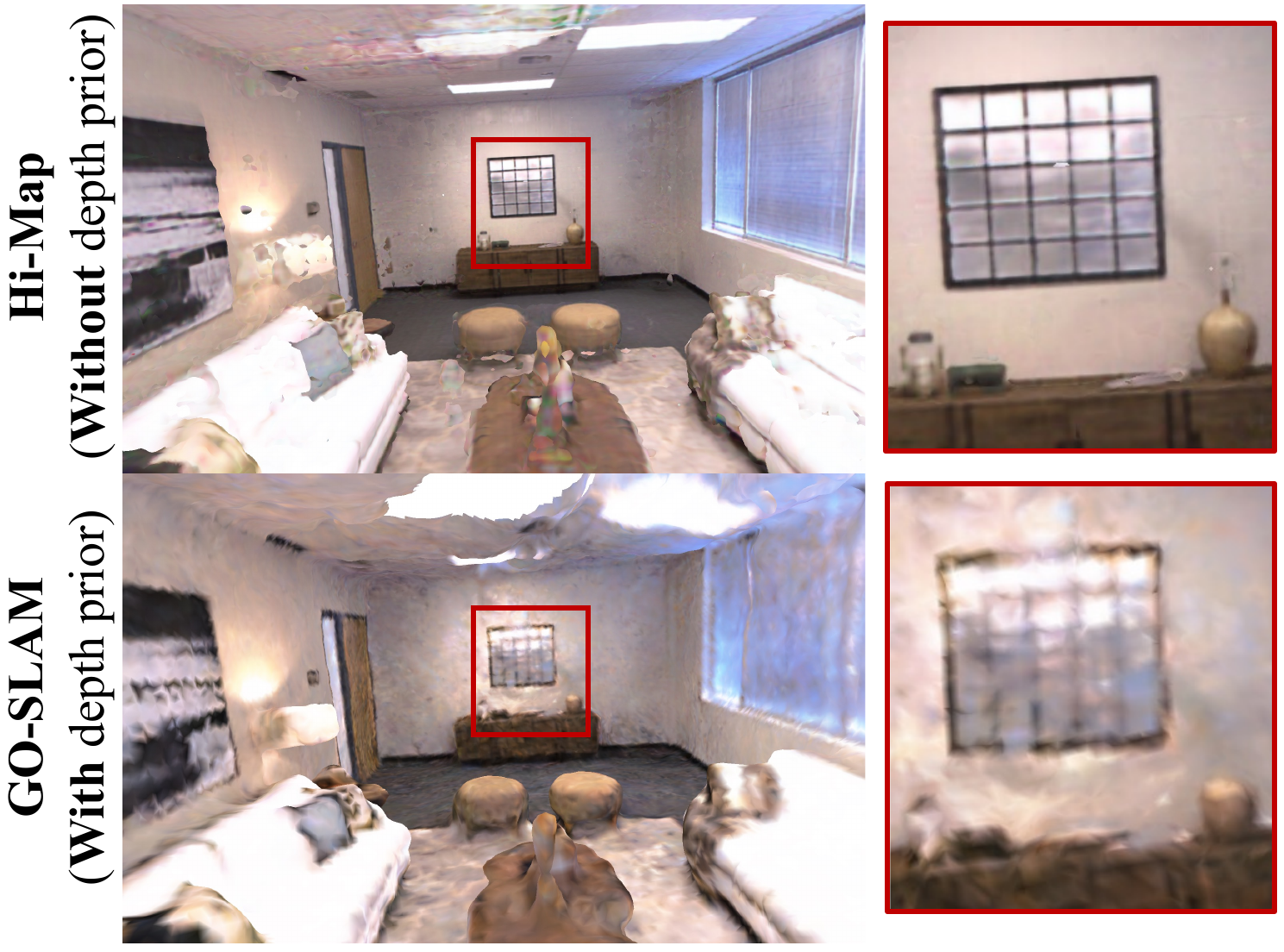}
    \vspace{-20pt}
      \caption{Our Hi-Map delivers higher mapping fidelity compared to existing state-of-the-art methods~\cite{goslam23} with monocular observations, even without the use of geometric priors derived from rigorous global optimization of external tracking systems.}
      \vspace{-20pt}
      \label{fig:cover}
\end{figure}

\begin{figure}[t!]
      \centering
      \vspace{-5pt}
      \includegraphics[scale=0.16]{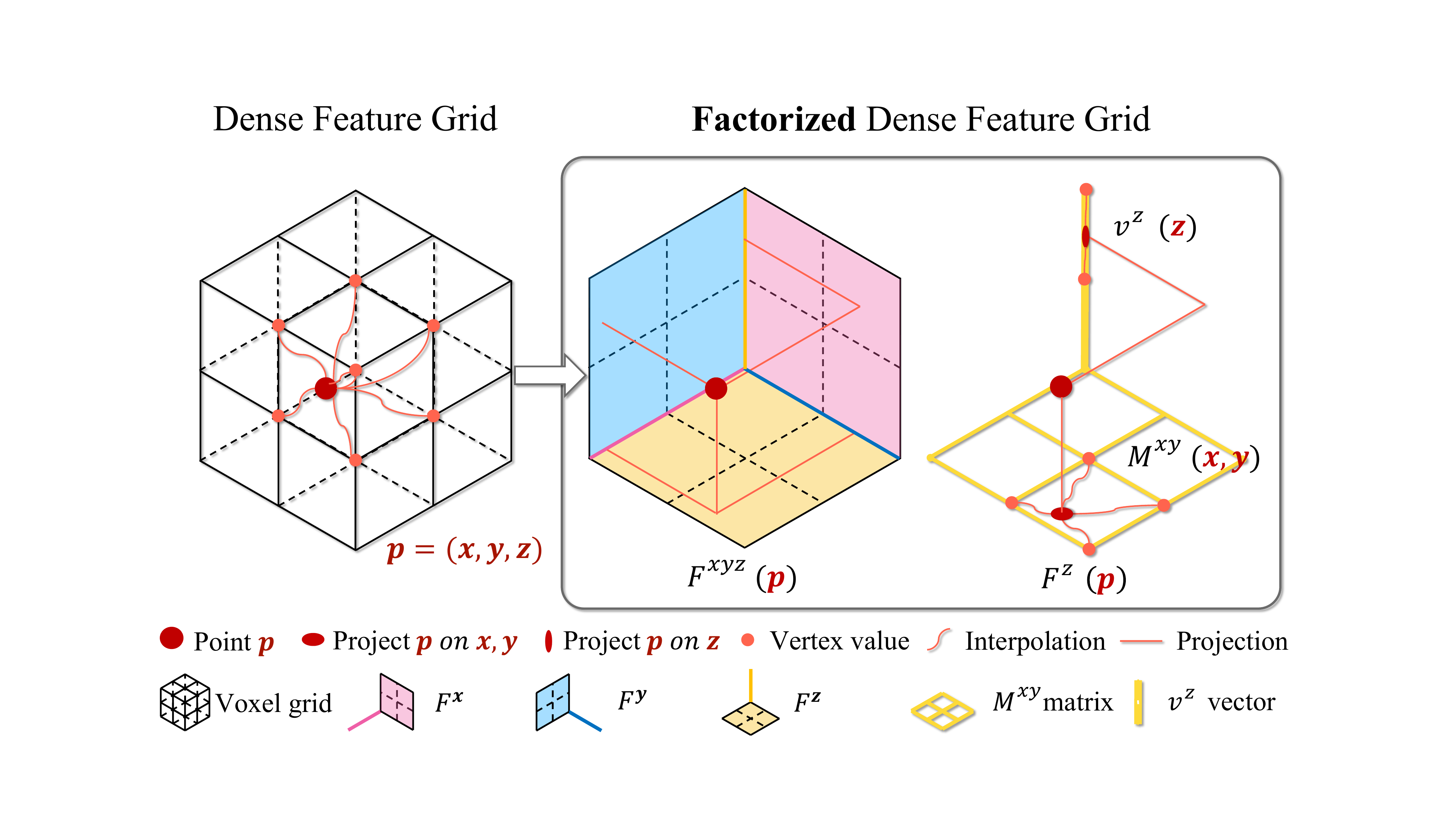}
      \caption{\textbf{Illustration of factorization scheme of a feature grid}. For a point $p$ of coordinate $(x,y,z)$, its value is assigned by performing trilinear interpolation at the 8 vertexes of the voxel when adopting dense feature grid encoding. When applying factorization, the value of $p$ is estimated by summing 3 components ($F^x, F^y, F^z$) to $F^{xyz}(p)$.
      An example is given for the value interpolation on component $F^z$, which includes the matrix component $M^{xy}$ and vector component $v^{z}$.}
      \vspace{-10pt}
      \label{fig:representation-F}
\end{figure}

In this paper, we introduce \textbf{Hi-Map}, a novel NeRF-based approach for efficient monocular dense mapping without relying on any depth priors.
To achieve this, we introduce a novel hierarchical representation by factorizing multi-resolution feature grids, inspired by~\cite{tensorf22}, where a low-rank regularization is proposed by factorizing the radiance field, leading to enhanced rendering quality and improved computation efficiency. 
This regularization technique simplifies the data structure, \ie, the 4D tensor, to lower-dimensional elements, namely low-rank components, to retain the most relevant feature for volume rendering.
Therefore, when extending to the context of dense mapping, such simplification, namely, factorization, can help retain the most relevant textural details in the RGB inputs for inferring the geometry, and thus facilitating faster convergence on novel views.
%
%
%
Specifically, we factorize the dense grid of each resolution level into separate orthogonal 2D planes and 1D lines, illustrated in~\cref{fig:representation-F}, where a coordinate is encoded no longer by grid vertexes but rather by planar and linear feature interpolations.
%
%
%
Expanding on the hierarchical factorized representation, we employ the Signed Distance Field (SDF) as a proxy to approximate volume density. By using this proxy-based approach for rendering, we capitalize on the benefits of SDF—namely, its coherent and accurate surface delineation—while circumventing the optimization instabilities it may cause.
%
%

Moreover, we introduce a dual-path encoding strategy to strengthen the photometric cues and further boost the reconstruction quality, especially for the distant and textureless regions.
%
%
Without depth priors, 
Hi-Map recovers view-independent geometry by incorporating absolute coordinates into the appearance encoding. 
We achieve this by allocating distinct factorized grids for geometry and appearance, where the appearance feature is combined with the samples' absolute coordinates. Such an encoding assists learning the variations in color and lighting caused by viewpoint shifts. On the other hand, overemphasis on such context in geometric features leads the representation to capture irrelevant textural correlations and thus degrades the reconstruction quality.

In summary, Hi-Map achieves efficient and high-fidelity dense mapping using solely posed RGB inputs and circumventing the need for external depth priors. Our contributions to this paper are as follows:

\begin{itemize}
    \item A novel hierarchical factorized representation for NeRF-based monocular dense mapping to achieve high-quality reconstructions without the need for any geometric priors.
    \item A dual-path encoding scheme effectively mitigates artifacts and enhances photometric consistency.
    \item A demonstrated superior performance on the Replica dataset~\cite{replica19arxiv} compared to the state-of-the-art monocular mapping methods~\cite{goslam23,imode23}, achieving about $50\%$ boost in incremental appearance and geometry estimation. For more details, please refer to our project homepage: \url{https://vlis2022.github.io/fmap/}.
\end{itemize}

\section{Related works}

Numerous methods for explicit dense mapping have been developed, primarily utilizing inputs from RGB-D sensors.~\cite{RGBDSLAM1,RGBDSLAM2,RGBDSLAM3,RGBDSLAM4,RGBDSLAM5,RGBDSLAM6,RGBDSLAM7}.
The Neural Radiance Field~\cite{nerf21}, a novel approach rooted in Implicit Neural Representation (INR) combined with volume rendering techniques, has inspired substantial implicit dense mapping~\cite{imap22,niceslam22,voxfusion22,eslam23,coslam23,nerfslam22,orbeez23,goslam23,newton23,hislam23arXiv}, resulting in higher reconstruction quality with more compact representation. Existing NeRF-based dense mapping can be generally divided into two categories based on its dependency on depth priors derived from sensors or estimations:

\begin{figure*}[t!]
      \centering
      \vspace{-5pt}
      \includegraphics[scale=0.26]{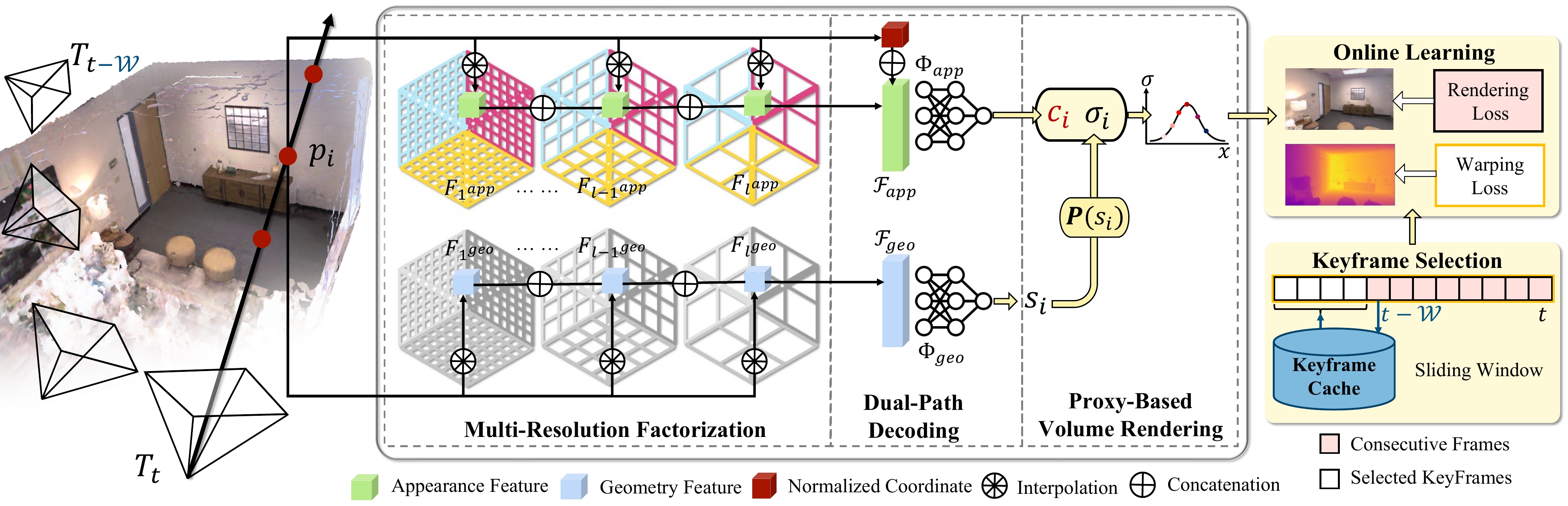}
      \caption{\textbf{The proposed pipeline of our Hi-Map}. Given a posed RGB frame $T_t$, the sampled coordinate $p_i$ is encoded by the Multi-resolution Factorized Feature Grid $F_l$ for appearance $\mathcal{F}_{app}$ and geometry $\mathcal{F}_{geo}$, which is decoded by $\Phi_{app}$ and $\Phi_{geo}$ to color ($c_i$) and SDF ($s_i$) through a Dual-Path Decoding, respectively. The volume rendering is performed based on the Proxy function $\boldsymbol{P(\cdot)}$ that transforms SDF to its density ($\sigma_i$), enabling continuous learning of neural implicit mapping on the observations in sliding window per timestep $t$.}
      \vspace{-10pt}
      \label{fig:Framework}
\end{figure*}

\noindent \textbf{Sensor Depth:}
The initial investigation by~\cite{imap22} revealed that a simple Multilayer Perceptron (MLP) is capable of functioning as a representation for incremental mapping that is trained online from scratch, by providing RGB-D camera inputs. This discovery stimulated further research to develop representations with improved performance on, \eg, scalability and computational efficiency~\cite{niceslam22,voxfusion22,eslam23,coslam23}. These implicit representations share the inherent ability of dense point cloud compression, spurring many following studies specifically tailored for robotics or automated driving scenarios~\cite{shine-map23,nerf-loam23,rim23,H2Mapping23}. Recently, some works went further to explore the larger-scale mapping or multi-robot mapping fusion~\cite{sbumap_map-fusion23,sbumap_mips23,sbumap_nisb23,cpslam23}, gradually bridging the NeRF-based mapping into the real-world application. 
However, depth sensors are not always available, which has prompted some studies to explore the possibility of NeRF-based mapping without reliance on sensor depth input.
\noindent \textbf{Estimated Depth:}
Attempts have been made to explore the monocular dense mapping that requires only RGB inputs. The immediate solution is to leverage off-the-shelf monocular depth estimation models~\cite{hislam23arXiv,nicer23}. Alternatively, the depth priors are provided by external SLAM systems, which provide globally consistent geometric cues~\cite{imode23,newton23,nerfslam22,orbeez23}. 
Such depth estimation can also be internalized by leveraging the multi-view stereo tactics to impose warping constraints~\cite {dimslam23}. 
However, these methods have generally struggled to achieve a balance between accuracy and computational efficiency, either by finding it hard to retain rendering fidelity~\cite{imode23,goslam23,orbeez23} or relying on external systems~\cite{hislam23arXiv,orbeez23,newton23,nerfslam22} and inefficient computation~\cite{nicer23,dimslam23}.
%

\section{The Proposed Hi-Map}
We present \textbf{Hi-Map}, a NeRF-based monocular dense mapping, specifically designed for incremental reconstruction independently of any depth priors, as the pipeline illustrated in~\cref{fig:Framework}. This system processes a stream of posed RGB inputs, leveraging hierarchical factorized grids and MLPs for scene representation, detailed in~\cref{subsec:Factorization}. High-quality mapping is achieved through a dual-path encoding strategy for geometry and appearance~\cref{subsec:Dual}, bolstered by a proxy-based volume rendering strategy as explained in~\cref{subsec:proxy}. Finally, the online optimization of mapping is detailed in~\cref{subsec:system}.

\subsection{Multi-Resolution Factorization}
\label{subsec:Factorization}

We aim to construct an implicit dense mapping that associates a spatial coordinate with its corresponding volume density and color, thereby enabling gradient-based volume rendering. We represent the scene with feature grids of multiple resolution levels and perform factorization on these feature grids, as depicted in~\cref{fig:representation-F}.
%

The dense feature grid can be viewed as a 4D tensor~\cite{tensorf22}, where each voxel is associated with latent features at their 8 vertices that represent either geometry or appearance. The factorization of the dense feature grid involves the 4D tensor decomposition.
%
For a dense grid $\mathcal{G}$ that assigns multiple feature channels to each voxel, representing the volume geometry and color, we define its factorization $\mathcal{F}$ as the sum of the subsequent 3 components $F^x$, $F^y$, and $F^z$ along grid axes x, y, and z, respectively:
{
\begin{equation}
\mathcal{F}
=\sum_{m\in xyz}F^m
=\mathbf{v}^x\circ\mathbf{M}^{yz}+\mathbf{v}^y\circ\mathbf{M}^{xz}+\mathbf{v}^z\circ\mathbf{M}^{xy}
\label{eq:fact}
\end{equation}}where $\mathbf{v}$ and $\mathbf{M}$ correspond to the line feature vector and plane feature matrix parts of component $F^m$. The $\circ$ symbol represents outer products. 
For a sample point $p$, its interpolated value in the factorized field is not computed by trilinear interpolation of the feature voxel as in $\mathcal{G}(p)$; instead, it is determined through bilinear interpolation ($BiLerp(\cdot)$) and linear interpolation ($LiLerp(\cdot)$) at the corresponding matrix and vector levels.
For example, in~\cref{fig:representation-F}, the interpolated value of $p=(x,y,z)$ at the $F^z$, which is the component resulting from decomposition along the z-axis, is calculated as:
{
\begin{equation}
\begin{gathered}
F^z(p)=
\mathbf{v}^z(z)\cdot \mathbf{M}^{xy}(x,y) \\
= LiLerp(z,\mathbf{v}^z) \cdot BiLerp(xy, \mathbf{M}^{xy})
\end{gathered}
\label{eq:bi-li}
\end{equation}}This operation reduces the memory footprint and computation that were originally required for storing the complete 4D tensors and performing interpolation among them.
Therefore, allocating grids of multiple resolutions has become a cost-effective strategy, enabling high-fidelity reconstruction of objects across a range of sizes and distances.

Considering the representation of a scene with multiple dense grids of different resolution levels $\mathcal{G}_l$, the total factorized feature volume would be:
{
\begin{equation}
\mathcal{F}
=\sum_{m\in xyz}F^m_1 \oplus \sum_{m\in xyz}F^m_2 \oplus ... \oplus \sum_{m\in xyz}F^m_L 
\label{eq:multi-fact}
\end{equation}}where $l \in (1,2, ..., L)$ represents the resolution levels, and $\oplus$ symbolizes the concatenation operation. 

\subsection{Dual-Path Decoding}
\label{subsec:Dual}

By assigning separate feature volumes to geometry and appearance, our approach ensures that each attribute is represented with an appropriate resolution and set of feature channels, resulting in a representation that is both specialized and adaptable.
In~\cref{fig:Framework}, a sampled coordinate $p_i$ is encoded through hierarchical factorized grids, yielding distinct feature representations for appearance and geometry at each resolution level. These feature representations are decoded to the Signed Distance Field (SDF), denoted as $s_i$, and color, denoted as $c_i$, by two separate small MLPs, \ie, $\Phi_{geo}$ and $\Phi_{app}$.
Notably, the geometric feature $\mathcal{F}_{geo}$ is directly decoded by $\Phi_{geo}$ to SDF:
{
\begin{equation}
s_i = \Phi_{geo}(\mathcal{F}_{geo}(p_i)),
\label{eq:geo-func}
\end{equation}}while the appearance feature $\mathcal{F}_{app}$ is combined with normalized spatial coordinate of $p_i$ before being decoded by $\Phi_{app}$:
{
\begin{equation}
c_i = \Phi_{app}(p_i , \mathcal{F}_{app}(p_i)).
\label{eq:app-func}
\end{equation}}Incorporating the coordinates of samples into our model provides global context, enabling a stable estimation of appearance regardless of viewing angle. Additionally, this method reinforces the coherence between geometry and color, guaranteeing a robust alignment between these attributes, despite their separate encoding in distinct feature volumes.

\subsection{Proxy-Based Volume Rendering}
\label{subsec:proxy}

\begin{figure}[t!]
      \centering
      \vspace{-5pt}
      \includegraphics[scale=0.415]{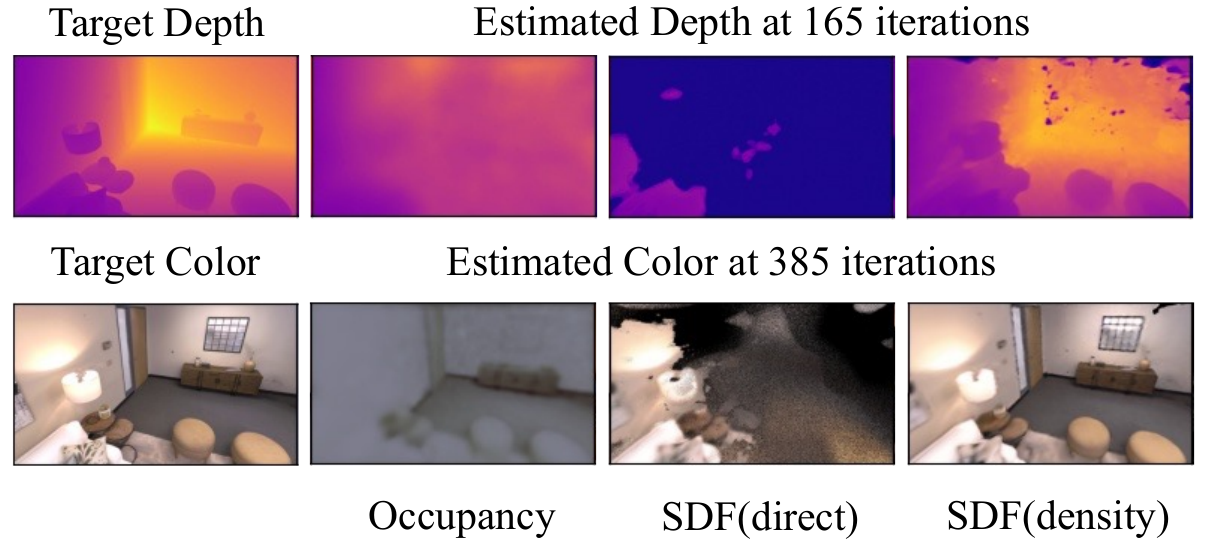}
      \caption{\textbf{Impact of geometric representations on volume rendering}. Hi-Map leverages SDF (density) representation, which includes a transformation of SDF to density. Consequently, it leads to a smoother gradient of weights compared to SDF (direct), where SDF is directly transformed into a weighting factor. SDF (density) also demonstrates faster convergence compared to occupancy.}
      \vspace{-10pt}
      \label{fig:sdf-occ}
\end{figure}

Unlike the direct transformation of SDF to weighting factor for color rendering, as suggested by many neural implicit mapping methods based on RGB-D~\cite{hislam23arXiv,H2Mapping23,neuralrgbd22}, we register the SDF as volume density with a proxy function $\boldsymbol{P}(\cdot)$, as depicted in~\cref{fig:Framework}, inspired by~\cite{eslam23,stylesdf22}:
{
\begin{equation}
\sigma_i= \boldsymbol{P}( s_i)
=\beta\cdotp sigmoid(-\beta\cdotp s_i),
 \label{eq:prox-func1}
\end{equation}}where $\beta$ is a trainable parameter. The inferred volume density, denoted as $\sigma_i$, is subsequently transformed into the final weighting factor, similar to the $\alpha-$composition:
{
\begin{equation}
w_i = exp(-\sum\limits_{k=1}^{i-1} \sigma_i) (1-exp(-\sigma_i)).
 \label{eq:prox-func2}
\end{equation}}where $w_i$ is the weighting factor for rendering a pixel by integrating the weighted samples along the corresponding camera ray. This rendering arrangement, denoted as SDF (density) in~\cref{fig:sdf-occ}, has demonstrated smoother and faster convergence compared to its occupancy-based and SDF (direct) alternatives.

\subsection{Mapping Optimization}
\label{subsec:system}
\noindent\textbf{Online Training: }Upon the arrival of new $n$ posed RGB frames, the system enables dense incremental reconstruction by minimizing the photometric rendering loss:
\begin{equation}
\mathcal{L}_c =\frac{1}{\mathcal{M}} \sum\limits_{x \in \mathcal{M}} \|{c_x - \tilde{c}_x}\|_1
  \label{eq:render_loss}
\end{equation}where $\mathcal{M}$ denotes the set of sampled pixels originating from the current sliding window, which stores a fixed number $\mathcal{W}$ of frames for optimization at any given time $t$. This is a common practice for managing computational resources and ensuring real-time performance, by not storing the entire sequence. The rendered color $\tilde{c}_x$ at pixel $x$ is formulated as the sum of the weighted colors along the ray, \ie, $\tilde{c}_x=\sum^N_{i=1} w_i\cdotp c_i$ following~\cref{eq:prox-func2}.
%
To achieve online convergence without any depth priors, an additional photometric warping constraint is included to best leverage the cross-frame photometric consistency, following the principles of depth estimation from multi-view stereo. For a patch $q_x$ centered around the pixel $x$, we utilized the multi-scale warping function $W(\cdot)$ proposed in~\cite{dimslam23}:
\begin{equation}
\mathcal{L}_{w} = \frac{1}{\mathcal{M}} 
\sum_{x \in \mathcal{M}} \sum_{l \in \mathcal{W}}
 SSIM(q_x, W_l(q_x,\tilde{d}))
  \label{eq:warp}
\end{equation}The Structural Similarity Index Measure ($SSIM$) is used to calculate the difference for the target patch $q_x$ to be warped to another frame $l$. The warping loss $\mathcal{L}_{w}$ is optimized by approximating the estimated pixel depth $\tilde{d}$ to the underlying true geometry. The depth is initialized by integrating volume density along the camera ray, \ie,$\tilde{d}_x=\sum^N_{i=1} w_i\cdotp z_i$,
%
where $z_i$ is the depth along the camera ray at point $p_i$. The total loss function is the summation of these components, weighted by factors $\alpha_{c}$ and $\alpha_{w}$:
\begin{equation}
\mathcal{L} = \alpha_{c}\mathcal{L}_{c} + \alpha_{w}\mathcal{L}_{w}
  \label{eq:final_loss}
\end{equation}
\noindent\textbf{Keyframe Selection:}
%
The implicit function is initialized over $N_{init}$ iterations and kicks off the mapping process that is updated by optimized for $N_{online}$ iterations upon every $n$ newly received observations. 
Throughout the incremental reconstruction process, a fixed number of frames is maintained within the active sliding window. This set includes $\mathcal{W}_{global}$ frames drawn from the global keyframe cache, as well as $\mathcal{W}_{local}$ consecutive frames preceding the current observation at time $t$, known as local frames, as depicted in Figure~\ref{fig:Framework}. 
The global keyframes are randomly sampled based on their overlap with the current observations, akin to the approach outlined in~\cite{niceslam22}.
Subsequent to each optimization, the earliest local frame among the removed set is added to the keyframe cache, with the other $n-1$ oldest local frames being removed from the sliding window.

\vspace{-3pt}



\begin{table*}[ht!] 
\footnotesize
  \centering
\begin{threeparttable}
  \centering
  \caption{Quantitative comparison of Hi-Map on Replica Dataset.}
  \label{tab:main}
   \setlength{\tabcolsep}{1.8mm}
  \centering
  \begin{tabular}{cc|cccccccc|c}
    \toprule
    Metrics &  Method&  Room 0&  Room 1& Room 2 & Office 0 & Office 1 & Office 2 & Office 3 & Office 4 & Avg.
    \\
    \midrule
    \multirow{2}{*}{PSNR $\uparrow$}
                                        &  GO-SLAM* &  14.30&  16.34&  17.43& 18.23& 20.79& 13.31& 14.07& 15.25 & 16.18\\
                                        &  \textbf{Hi-Map} &  \textbf{23.48}& \textbf{27.81}& \textbf{27.09}& \textbf{32.65}& \textbf{33.74}& \textbf{24.23}& \textbf{22.72}& \textbf{27.15} & \textbf{27.36}\\
    \cmidrule(lr){2-11}
    \multirow{2}{*}{SSIM $\uparrow$} 
                                    &  GO-SLAM* &  0.37& 0.47& 0.49& 0.38& 0.44& 0.49& 0.47& 0.51 & 0.45\\
                                    &  \textbf{Hi-Map}&   \textbf{0.70}& \textbf{0.78}& \textbf{0.81}& \textbf{0.86}& \textbf{0.85}& \textbf{0.78}& \textbf{0.75}& \textbf{0.84} & \textbf{0.80}\\
    \cmidrule(lr){2-11}
    \multirow{2}{*}{Depth L1 $\downarrow$}
                                          &  GO-SLAM* &  0.33& 0.24& 0.27& 0.20& 0.18& 0.31& 0.47& 0.36 & 0.30\\
                                          &  \textbf{Hi-Map} &  \textbf{0.15}& \textbf{0.04}& \textbf{0.11}& \textbf{0.03}& \textbf{0.02}& \textbf{0.17}& \textbf{0.38}& \textbf{0.17} & \textbf{0.13}\\
    \midrule
    \multirow{3}{*}{Acc. $[cm]\downarrow$}& iMODE~\cite{imode23} &  7.40& 6.40& 9.30& 6.60& 11.80& 11.40& 9.40& 8.00& 8.78\\
                                          &  GO-SLAM* &  5.58& 4.68& -& 3.27& 4.09& 4.76& 5.21& 4.70& 4.61\\
                                          &  \textbf{Hi-Map}  &  \underline{6.51}& \underline{4.93}& \textbf{5.10}& \underline{3.55}& \textbf{3.45}& \underline{7.06}& \underline{9.50}& \underline{7.70} & \underline{5.98}\\
    \cmidrule(lr){2-11}
    \multirow{3}{*}{Comp. $[cm]\downarrow$}&  iMODE~\cite{imode23} &  13.50& 10.10& 19.20& 9.70& 17.00& 14.50& 11.80& 15.40& 13.90\\
                                           &  GO-SLAM* &  9.12& 7.43& -& 13.17& 13.60& 10.79& 9.28& 9.13 & 10.36\\
                                           &  \textbf{Hi-Map}  &  \textbf{6.10}& \textbf{5.25}& \textbf{6.01}& \textbf{11.60}& \textbf{10.49}& \textbf{6.89}& \textbf{6.62}& \textbf{6.36}& \textbf{7.42}\\
    \cmidrule(lr){2-11}
    
    \multirow{3}{*}{Comp. Ratio$[\%]\uparrow$}&  iMODE~\cite{imode23} &  38.70& 46.10& 36.10& 49.3& 30.10& 29.80& 36.00& 31.00 & 37.10\\
                                              &  GO-SLAM* &  59.10& 59.19& -& 65.08& 59.73& 58.95& 53.60& 56.48& 58.88\\
                                              &  \textbf{Hi-Map}  &  \textbf{75.91}& \textbf{70.78}& \textbf{71.42}& \textbf{76.04}& \textbf{72.84}& \textbf{68.01}& \textbf{65.34}& \textbf{70.77}& \textbf{71.39}\\
 \bottomrule
  \end{tabular}
\begin{tablenotes}
\item 
The '-' symbol indicates the failure case that was validated 5 times and is not included in the calculation of the average value.
The '*' symbol indicates the results are obtained from its official open-source implementation for GO-SLAM~\cite{goslam23} and evaluated using the same evaluation pipeline as our method. The Depth L1, PSNR, and SSIM are evaluated at the last iteration of every mapping optimization. To facilitate fare comparison, the depth L1 of GO-SLAM is aligned with the ground-truth depth using the median value, as its provided pose stream shares the scale ambiguity.

\end{tablenotes}
\end{threeparttable}
  \vspace{-5pt}
\end{table*}

\begin{figure*}[t!]
      \centering
      \includegraphics[scale=0.45]{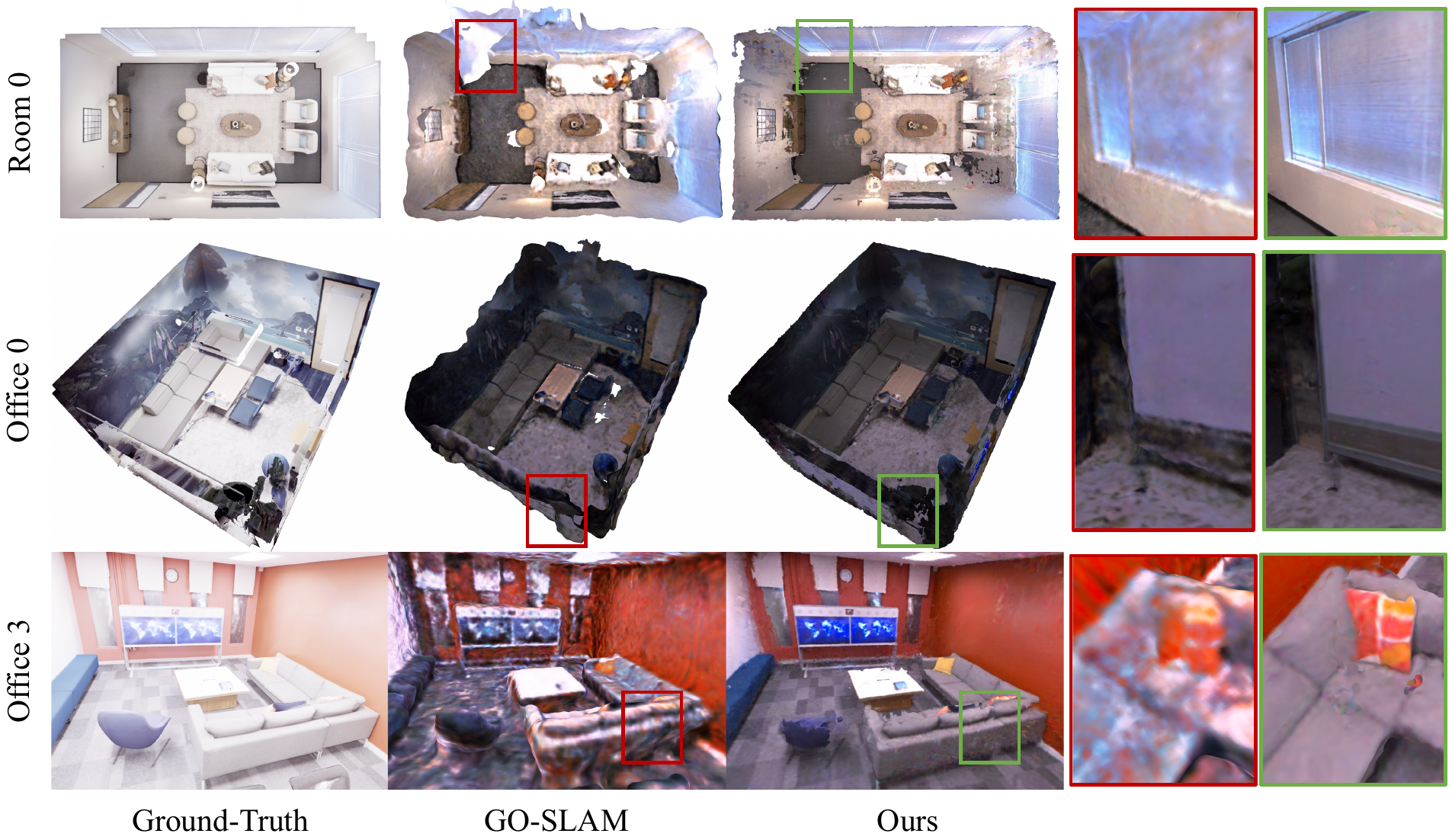}
      \caption{\textbf{Comparison of final reconstruction on Replica dataset}. The blind spot regions are delineated with \textbf{\color{darkred}red (GO-SLAM)} and \textbf{\color{darkgreen}green (Hi-Map)} boxes, respectively, and corresponding visualizations are provided from observable viewpoints. Our approach achieves higher scene fidelity and exhibits stronger expressive capability for indoor vertical planes.}
      \vspace{-10pt}
      \label{fig:replica}
\end{figure*}

\begin{figure}[t!]
      \centering
      \vspace{-5pt}
      \includegraphics[scale=0.2]{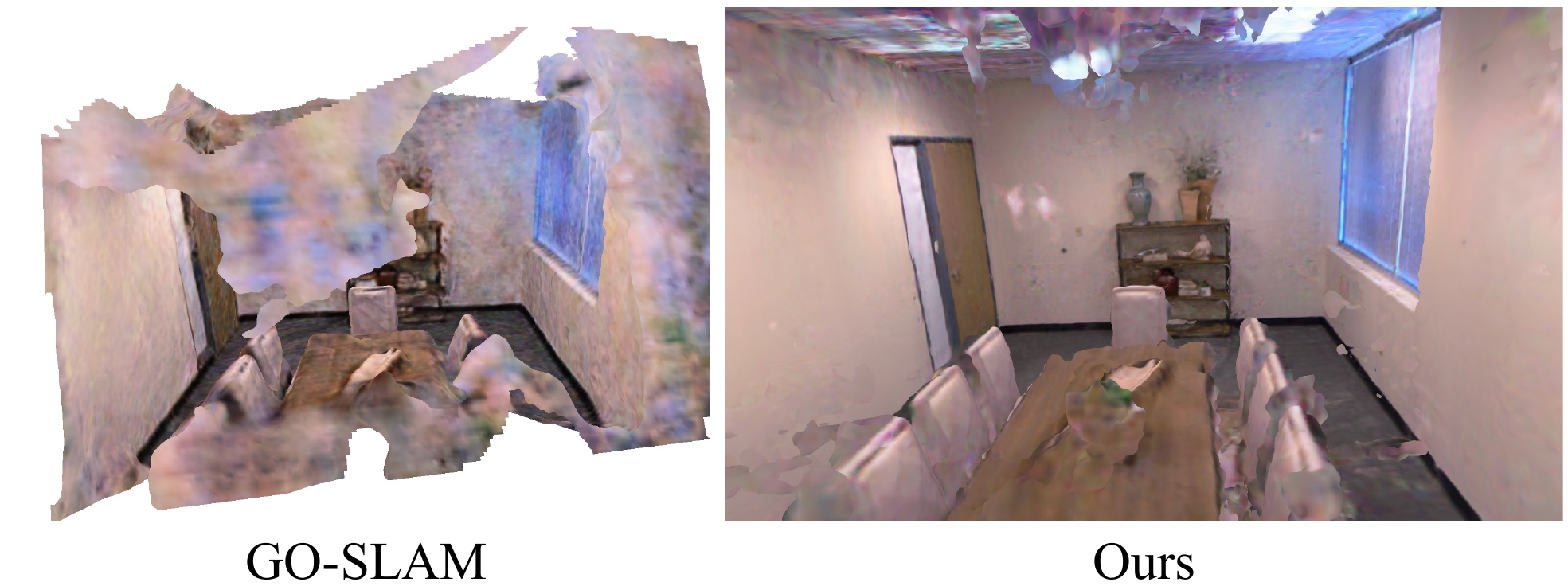}
      \caption{\textbf{Demonstration of the risk of relying on unreliable depth prior}. Reconstruction of Room 2 sequence where GO-SLAM failed to reconstruct the whole scene.}
      \vspace{-15pt}
      \label{fig:fail-room2}
\end{figure}

\section{Experiments}
\vspace{-3pt}
We evaluate our method on the Replica dataset~\cite{replica19arxiv} and TUM dataset~\cite{tumrgbd12}, comparing it with other state-of-the-art monocular dense mapping frameworks~\cite{goslam23,imode23}.

\subsection{Experimental Settings}

\noindent \textbf{Implementation Details: }
We conducted experiments using a 2.10GHz Intel Xeon Gold 5218R CPU and an NVIDIA GeForce RTX 3090, and 2.60GHz Intel Xeon Platinum 8358P CPU, and an A800-SXM4-80GB GPU.
Our mapping framework is initialized on $\mathcal{W}_{init}=15$ frames for $N_{init}=1500$ iterations, where the color gradient is back propagated until $N_{c}=250$ iterations. During the continuous mapping process, we maintain a sliding window of $\mathcal{W}=20$ frames, where $\mathcal{W}_{global}=5$ and $\mathcal{W}_{local}=15$. $N_{online}=20$ iterations of optimization are performed to update the map for every $n=5$ frame. The oldest $n$ local frames are removed while the new $n$ incoming frames are added to the window for the next map update.
The feature grid resolution and vertex feature channels are set differently for geometry and appearance encoding. Both are limited by a coarsest resolution of 64cm and a finest resolution of 2cm. For geometric encoding, the grid resolution of 6 layers is evenly spaced between 2cm and 64cm, with 2 feature channels per level. For appearance encoding, we use coarse and fine feature spatial divisions with resolutions of 24cm and 2cm, while increasing the feature channels to 32. The features of each resolution level are combined for processing by the corresponding geometry and appearance decoders, which consist of shallow MLPs with 2 layers and 32 hidden channels.
We use the Adam optimizer with learning rates set to 0.01 and 0.00001 for the factorized grid features and MLP decoders, respectively. We configured the rendering loss as $\lambda_c=0.1$ during initialization and $\lambda_c=0.001$ during online mapping, with the warping loss set at $\lambda_w=1.0$.

\noindent \textbf{Evaluation Metrics: }
We assess the quality of reconstruction using three well-established metrics: Accuracy (Acc.[cm]), Completion (Comp.[cm]), and Completion Ratio (Comp.[$\%$]), which measures the proportion within a 5cm threshold. In contrast to static 3D reconstruction tasks, incremental mapping places additional emphasis on continuous estimation performance. 
Therefore, we additionally assess the Structural Similarity Index (SSIM), Peak Signal-to-Noise Ratio (PSNR[db]), and the L1 term of the estimated depth (Depth L1[cm]), calculated after completing a mapping update and compared the average values across the complete sequence.
Such performance evaluation for continuous mapping is carried out on the methodologies that are fully open-source at the time of submission.

\subsection{Evaluation of Mapping}
\begin{figure}[h!]
      \centering
      \includegraphics[scale=0.2]{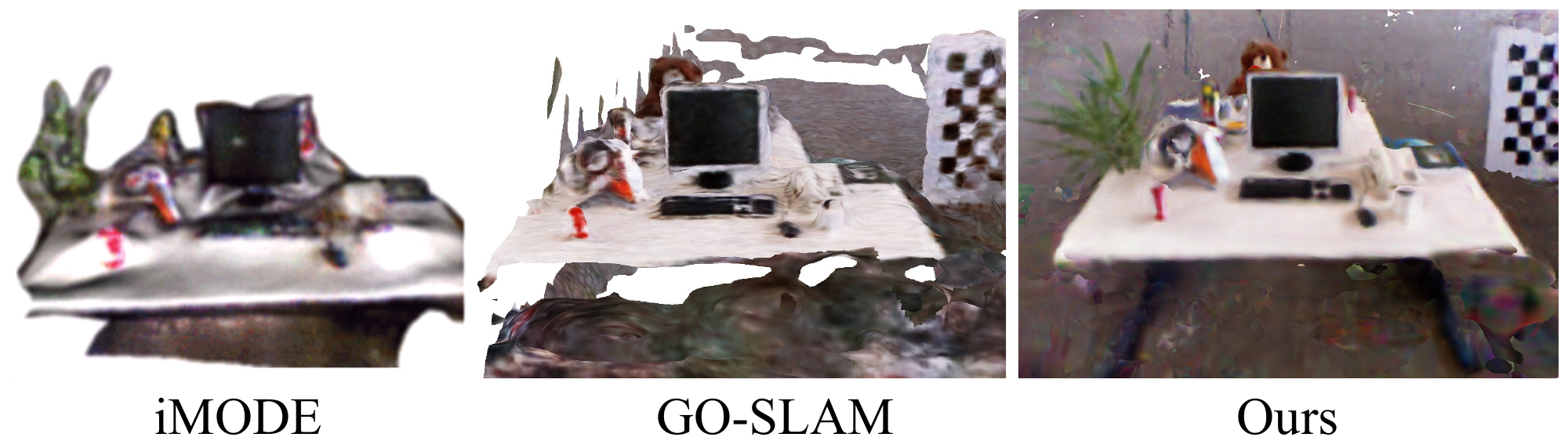}
      \caption{\textbf{Comparison of reconstruction on TUM RGBD dataset}. The visualization of iMODE is directly retrieved from the original manuscript~\cite{imode23}.}
      \vspace{-10pt}
      \label{fig:tum}
\end{figure}

We first evaluate the Hi-Map quantitatively in~\cref{tab:main}. Our method demonstrates overall higher rendering quality compared to GO-SLAM~\cite{goslam23} throughout the entire process, evaluated by SSIM, PSNR, and Depth L1. 
The final reconstruction metrics show that our method produces the most complete reconstruction. However, the high completion can compromise the overall reconstruction quality, because the regions where our baselines fail to complete are barely observable, which increases the difficulty for highly accurate estimation. Nevertheless, we achieved a secondary ranking on average accuracy even when the GO-SLAM failed at the Room 2 sequence and thus exonerated from the calculation. Such failure also indicates that the reliance on provided geometric cues from vSLAM, as depicted in~\cref{fig:fail-room2}, could be unreliable, diminishing the robustness of overall reconstruction.
The qualitative comparison is available in~\cref{fig:replica}, demonstrating our capability of online high-fidelity reconstruction. Notably, our method can generate large and smooth structures while maintaining the expressiveness of the authentic details, thanks to the inherent planer feature cues and the multi-level encoding. The visualization of results on the TUM RGBD dataset also supports the performance of our method, by demonstrating a more complete and detailed reconstruction in~\cref{fig:tum}.

\subsection{Ablation Study}

\begin{figure}[t!]
      \centering
      \includegraphics[scale=0.5]{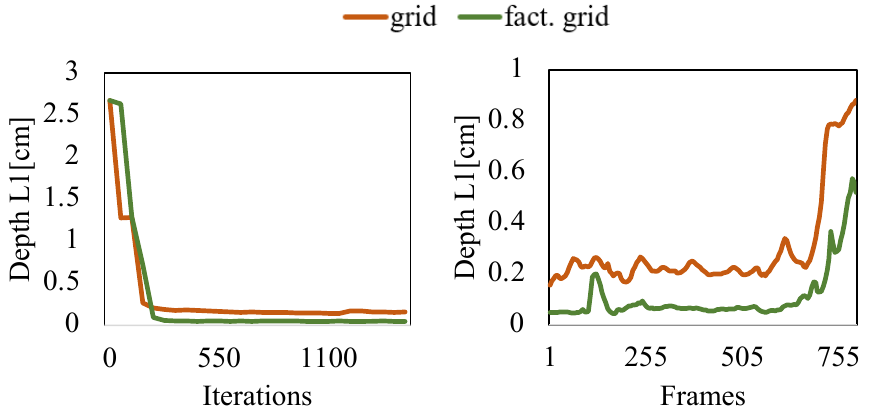}
      \caption{\textbf{Ablation of factorization}. The depth L1 loss is consistently smaller for both initialization (left) and continuous (right) mapping stages when incorporating the factorization schemes.}
      \label{fig:ablation-fact}
\end{figure}

\begin{figure}[t!]
      \centering
      \vspace{-5pt}
      \includegraphics[scale=0.5]{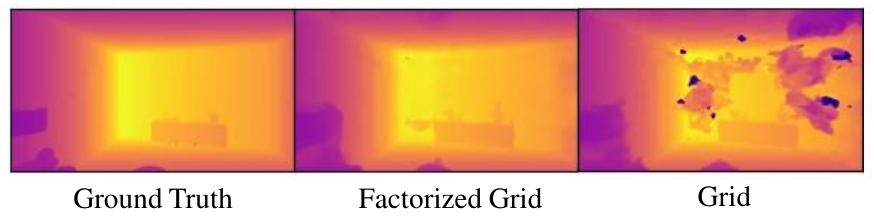}
      \caption{\textbf{Ablation of factorization}. Visual demonstration of ablation.}
      \label{fig:ablation-fact-fig}
      \vspace{-15pt}
\end{figure}

\noindent\textbf{Factorization:} The introduction of Low-rank regularization to the feature grid optimization, \ie, factorization, leads the representation to smoothly generalize to new observations, as shown in~\cref{fig:ablation-fact-fig}. 
Such a factorization scheme tends to simplify the representation by removing the less impactful features in, \eg, textureless region, which creates large artifacts in the optimization of grid representation. 
Therefore, we can effectively capture the underlying structure of the scene, contributing to higher-quality output.
Such effectiveness is also supported by numerical evidence in~\cref{fig:ablation-fact}, where factorized grid structure enables consistently superior geometric rendering throughout the mapping process. 

\begin{figure}[h!]
      \centering
      \vspace{-5pt}
      \includegraphics[scale=0.4]{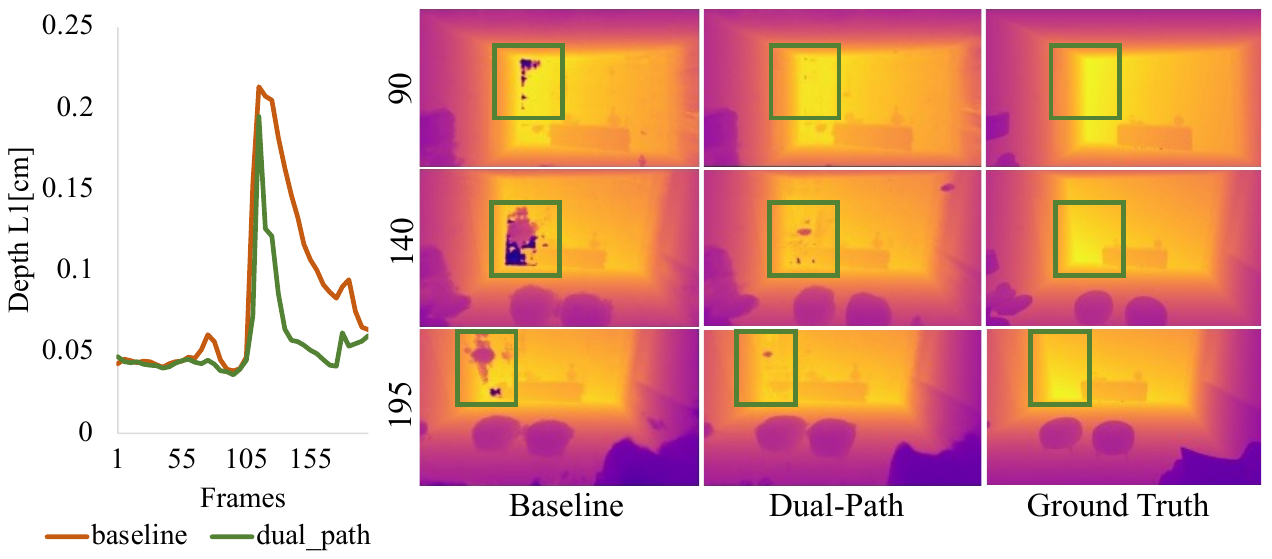}
      \caption{\textbf{Ablation of dual-path encoding.} The geometric consistency and estimation accuracy are significantly booted when the appearance feature is jointly encoded with absolute coordinates. }
      \vspace{-10pt}
      \label{fig:ablation-dual-path}
\end{figure}

\noindent\textbf{Dual-Path Encoding }
enhanced the geometric consistency of the feature encoding, demonstrated in~\cref{fig:ablation-dual-path}. Without this encoding strategy, the geometry in textureless areas could not be accurately reconstructed within limited optimization iterations. 
The reason is that the absence of distinct textures in these regions creates ambiguity when establishing cross-frame warping constraints thus leading to the loss of geometric details, which are recovered by implicitly learning from the coordinate-associated appearance features.


\section{CONCLUSIONS}
In this paper, we have presented Hi-Map for monocular dense mapping. By uniquely integrating a hierarchical factorized grid with a dual-path encoding strategy, Hi-Map achieved high-fidelity 3D reconstruction using only posed RGB inputs, without the need for external depth priors. Our method not only enhanced memory efficiency and mapping quality but also significantly improved reconstruction in challenging areas such as remote and textureless regions, achieving overall higher geometric and textural accuracy compared to the existing state-of-the-art methods.
\bibliographystyle{./IEEEtran}
\bibliography{./IEEEabrv,./ref}
\end{document}